\documentclass[sigconf]{acmart}

\usepackage{booktabs} 
\usepackage{mathrsfs}
\usepackage{multirow}
\usepackage{url}
\usepackage{balance}

\copyrightyear{2019} 
\acmYear{2019} 
\setcopyright{acmcopyright}
\acmConference[ICMI '19]{2019 International Conference on Multimodal Interaction}{October 16--20, 2018}{Suzhou, Jiangsu, China}
\acmBooktitle{2018 International Conference on Multimodal Interaction (ICMI '19), October 16--20, 2018, Suzhou, Jiangsu, China}
\acmPrice{15.00}
\acmDOI{10.1145/3242969.3264981}
\acmISBN{978-1-4503-5692-3/18/10}
\begin{document}
\title[Bootstrap Model Ensemble and Rank Loss for Engagement Intensity Regression ...]{Bootstrap Model Ensemble and Rank Loss for Engagement Intensity Regression}

\author{Kai Wang}
\affiliation{%
	\institution{SIAT\authornote{SIAT: Shenzhen Institutes of Advanced
			Technology, Chinese Academy of
			Sciences}, 
		Shenzhen Key Laboratory of Virtual Reality and Human Interaction Technology}
	\city{P.R. China}
}

\author{Jianfei Yang}
\authornote{equal contribution with Kai Wang}
\affiliation{%
  \institution{School of Electrical and Electronic Engineering\\
  	Nanyang Technological University}
  \city{Singapore}
}

\author{Da Guo}
\affiliation{%
	\institution{SIAT\authornote{SIAT: Shenzhen Institutes of Advanced
			Technology, Chinese Academy of
			Sciences}, 
		Shenzhen Key Laboratory of Virtual Reality and Human Interaction Technology}
	\city{P.R. China}
}

\author{Kaipeng Zhang}
\affiliation{%
  \institution{The University of Tokyo}
  \city{Japan}
}

\author{Xiaojiang Peng}
\authornote{corresponding author}
\affiliation{%
	\city{P.R. China}
}

\author{Yu Qiao}
\authornote{common corresponding author}
\affiliation{%
	\institution{SIAT, 
	Shenzhen Key Laboratory of Virtual Reality and Human Interaction Technology}
	\city{P.R. China}
}


\begin{abstract}
This paper presents our approach for the engagement intensity regression task of EmotiW 2019. The task is to predict the engagement intensity value of a student when he or she is watching an online MOOCs video in various conditions. Based on our winner solution last year, we mainly explore head features and body features with a bootstrap strategy and two novel loss functions in this paper. We maintain the framework of multi-instance learning with long short-term memory (LSTM) network, and make three contributions. First, besides of the gaze and head pose features, we explore facial landmark features in our framework. Second, inspired by the fact that engagement intensity can be ranked in values, we design a rank loss as a regularization which enforces a distance margin between the features of distant category pairs and adjacent category pairs. Third, we use the classical bootstrap aggregation method to perform model ensemble which randomly samples a certain training data by several times and then averages the model predictions. We evaluate the performance of our method and discuss the influence of each part on the validation dataset. Our methods finally win 3rd place with MSE of 0.0626 on the testing set. 
\end{abstract}

%
%

\ccsdesc[500]{Computing methodologies~Activity recognition and understanding}

\keywords{Engagement intensity prediction; multiple instance learning}

\maketitle

\section{Introduction}
Online education provides superior education resources and convenient access for students. In the online-learning environment such as Massive Open Online Courses (MOOCs), students can easily choose the courses they are interested in. In the regular courses where students are engaged with the class materials in a structured and monitored way, the instructors can directly observe students behavior and obtain feedback. Whereas, the distant nature and the sheer size of an online course require new approaches for providing student feedback and guiding instructor intervention. Therefore, building a robust student engagement intensity regression system is necessary and challenging.

In this challenge, we focus on automatic student engagement intensity regression using MOOC video data. The goal of such a regression model is to minimize the Mean Square Error (MSE) loss between the predicted and true engagement intensity. The task formulates the engagement into four levels ($0, 0.33, 0.66, 1$) indicating \textit{completely disengaged, barely engaged, engaged} and \textit{highly engaged}, respectively \cite{mustafa2018prediction}. In the official provided videos, the body and emotion of the student are the most useful content, but the data distribution of the official provided videos is unbalanced, which means that most of the videos are \textit{engaged}. Based on our EmotiW2018 submission and experiments, we find the performance is sensitive to different training and validation data splits. Then we propose a bootstrap model ensemble with three different data splits. Specifically, we randomly sample a certain training data by several times and then average the model predictions as our bootstrap method. Inspired by the fact that engagement intensity can be ranked in values, we design a rank loss as a regularization which enforces a distance margin between the features of distant category pairs and adjacent category pairs. Our regression framework consists of four parts, namely feature extraction, regression network (LSTM +FC), modality consensus and model ensemble. We extract three types of features, namely C3D, Openface, and Openpose(with 70 face landmarks). Then, the features are fed into regression network to predict the engagement intensity value. Finally, the modality consensus and model ensemble boost each single regression model.

\begin{figure*}[htp]
	\includegraphics[width=0.8\textwidth]{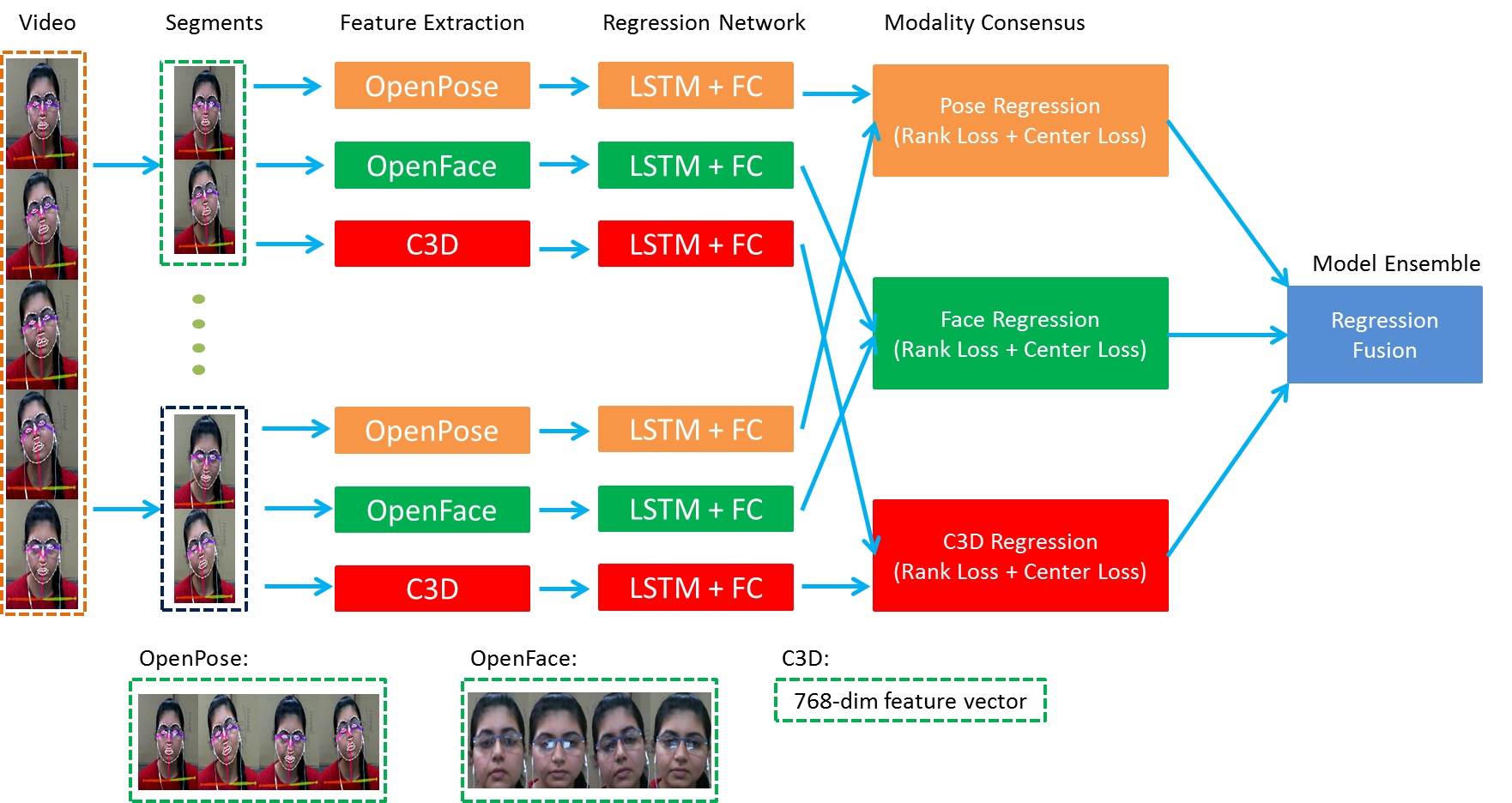}
	\caption{The system pipeline of our approach. }
	\label{fig:pipeline}
\end{figure*}

\section{Related Work.}
\textbf{Student Engagement.}
Student engagement is known to be a significant factor in the success of student learning \cite{8355953}, but there are still many challenges in robust engagement prediction. Many previous studies on automatic engagement recognition are focused on the perceived engagement (i.e., engagement received by an external people)\cite{gray2016effects}. Various automatic engagement prediction systems have been proposed using multi-modal information such as student response \cite{koedinger1997intelligent}, facial \cite{wang2018deep, wang2019region, tan2017group, meng2019frame} or body movements in learning videos \cite{d2010multimodal, xiao2017dynamics}, behavior in test quizzes \cite{joseph2005engagement} and even advanced physiological and neurological measures \cite{goldberg2011predicting}. Among them, the video data is a good trade-off between capturing convenience and granularity. Using videos, \citeauthor{whitehill2014faces} \cite{whitehill2014faces} analyze facial features and build a SVM classifier for engagement prediction. Apart from faces of subjects, their postures also play an important role in engagement, which is utilized in \cite{bosch2016detecting}. Another interesting work employs both facial features and test logs to analyze their learning levels\cite{d2009multimethod}. In EmotiW2018, some multi-instance learning based methods \cite{niu2018automatic, yang2018deep} have achieved significant improvements, which inspires us to design a multi-instance learning framework for engagement intensity regression.

\section{Approach}
\subsection{System Pipeline}
Figure \ref{fig:pipeline} shows the proposed system. To begin with, we divide the video into several segments and extract various features such as Convolutional 3D (C3D), Openface and OpenPose features. Second, we send the various features into the regression network (LSTM + FC). The regression network outputs the engagement intensity regression values of the segments. Third, the modality consensus is used to aggregate the values (average the segments' outputs). Finally, we ensemble the different modalities results as the video engagement intensity value. The regression network is optimized by Rank Loss (RL) and MSE loss. Furthermore, inspired by the bootstrap method, we resample the dataset for three times as the empirical distributions to estimate the true distribution of the dataset.

\subsection{Multiple Instance Learning framework}
Most of the previous works extract the multi-modal features directly from the whole video, which usually causes loss of segment detail information and leads to redundancy. Considering the weaknesses above, we formulate the student engagement as a multi-instance regression. A video sequence $V$ is divided into $k$ segments, i.e. $V = [s_1, s_2, s_3, ..., s_k]$, where $s_i$ represents the $i$-th video segment, and each video segment is regarded as an instance with the same label as the whole video. We obtain $M$ different modality features $F_k=[f_k^1, f_k^2, f_k^3, ..., f_k^m]$ from a segment and feed them into our framework. Section \ref{sec:feature} will introduce our features in detail. These features are fed into a LSTM and three fully connected layers. Note that regression parameters for different feature modality are different, while those of the same modality can adopt shared weights. Then we can obtain a regression value $r_i$ for each segment $s_i$. We average all the segment regression scores as the final prediction engagement intensity of the video, and optimize our network by RL and MSE losses between annotated engagement level and regression result. 

The training and validation video data are captured at 30fps and we extract 5 frames per second with a resolution of $640\times 480$. As the video length is different, we divide the video into several segments by averaging length. These video segments of one video sequence are the inputs of our proposed framework. The next step is to extract multi-modal features using different tools or pretrained models.

\subsection{Multi-modal Features}\label{sec:feature}
According to the foregoing, we should employ the modalities that are the most related factors of engagement. Imagine that in the video where a subject is watching an online educational clip, the movements of the subject should indicate their degree of concentration. Facial and postural changes reflect the degree of movement and engagement intensity. In addition, the specific actions such as writing, thinking or gazing can also help us predict the engagement intensity. Inspired by these factors and intuitions, we extract our features from two perspectives including three modalities (OpenFace, OpenPose and C3D features).

\subsubsection{OpenFace and OpenPose Feature: Gaze, Head and Body Posture}
It is obvious that engagement intensity is negatively correlated to the degree of human motions \cite{fredricks2004school}. If a student focus on the lecture content, the movements of gaze and pose may be very little. If the subject is barely engaged, he should be absent-minded with drifting gazes and substantial body movements. We capture the gaze and head movement features using OpenFace \cite{amos2016openface} while body posture characteristics via OpenPose \cite{simon2017hand}. 

\textbf{Eye Gaze}: OpenFace offers eye-gaze estimation and tracking data, and using it we obtain gaze coordinates for each face. We then calculate the average variance of the points compared to the mean location in each segment, and 6 gaze-related features are obtained for one segment.

\textbf{Head Pose}: OpenFace is able to extract head pose (translation and orientation) information in addition to facial landmark detection. Head movements are sometimes aligned with gaze drifting but sometimes not. This insight reminds us that head pose features should be complementary with gaze, and using both head and gaze may enrich the features

\textbf{Body Posture}: For further considerations, human body movement is also an indicator of engagement intensity. Their actions reflected by body movement should contain more information of specific purposes such as contemplation or writing notes. These actions can be captured by detection of body key points via OpenPose. We select 14 frequent-detected keypoints that indicates the upper body movement and also use their standard deviations as features. Since OpenFace and OpenPose work in different scheme, the information of body and face cannot be synchronized. We regard the body feature as an independent modality.

Although the OpenFace and OpenPose features include face, head and body, it is rather limited. These features can only represent the degree of movement of different component, but the concrete actions and gaze changing patterns are neglected. More severely, high-level extraction using open library loses more spatial features especially for faces in the video. Therefore, the action features need to be extracted to enrich the features.


\begin{figure*}[htp]
	\includegraphics[width=0.9\textwidth]{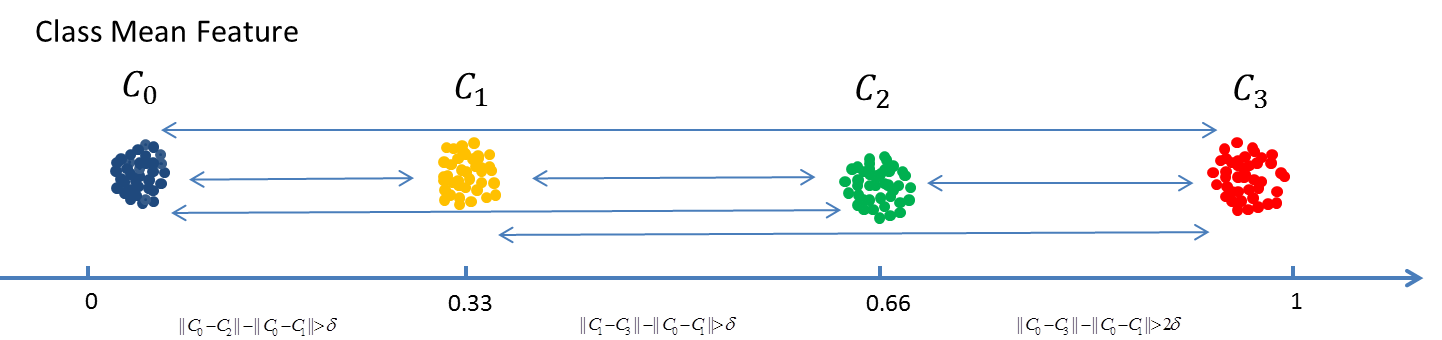}
	\caption{C\_i is the feature center of $i$th engagement intensity level. $\delta$ is the margin of different engagement  levels.}
	\label{fig:taks_2_loss}
\end{figure*}

\subsubsection{Action Features: C3D}
Recently, deep learning improves many computer vision tasks significantly. C3D\cite{tran2015learning, li2016online} is a novel deep spatio-temporal feature that achieves good performance in activity recognition. As we have modeled several type features of facial characteristics, C3D can be a robust representation for the body action in the spatio-temporal domain. Specifically, we crop the body by using the body landamarks offered by OpenPose, then use C3D pretrained model (in Sports-1M dataset) to extract the C3D feature in each segment. The body image is resized to $228\times228$ for C3D input. We finally obtain 768 dimensions of the C3D features as a modality for one segment.

\subsection{ MSE and Rank Losses}
We calculate the MSE and Rank losses between the ground truth and our predict engagement intensity. In the following paragraghs, we will introduce the MSE and our RL in detail. The MSE loss is very common defined as follows:
\begin{equation}
L_m = \frac{1}{n}\sum_{i=1}^{n}(Y_i-\frac{1}{k}\sum_{j=1}^{k}G_m(f_j^m))^2,
\end{equation}
where $G_m$ denotes the regression network transformation for feature $f_m$. This loss can optimize our framework for one modality $m$ and we can train the network for multiple modalities, obtaining $M$ regression parameters for $M$ modalities. 

Our rank loss can be summarized as follows:
\begin{align}
d_i^1 &= ||C_i - C_{i+1}||, i = {0,1,2} \\
d_i^2 &= ||C_i - C_{i+2}||, i = {0,1} \\
d_i^3 &= ||C_i - C_{i+3}||, i = {0} \\
\end{align}
where $d_i^1$, $d_i^2$, $d_i^3$ are the $L_2$ distances between the centers ($C_i$) of each engagement intensity level in feature space. The intuition is that $d_i^1$ should be smaller than $d_i^2$ and $d_i^2$ should be smaller than $d_i^3$. Inspired by Center Loss \cite{Wen2016A}, the intra-class $L_2$ distance should be as small as possible to make more discriminatory between the intensity levels.


\begin{equation}
    L_{rank1} = \sum_{i=0}^{1}\sum_{j=0}^{2}max({0, \delta - (d_i^2 - d_j^1))}
\end{equation}

\begin{equation}
    L_{rank2} = \sum_{i=0}^{0}\sum_{j=0}^{2}max({0, 2\delta - (d_i^3 - d_j^1))}
\end{equation}

\begin{equation}
    L_{CRL} =  L_{center} +  L_{rank1} + L_{rank2}
\end{equation}

Where $L_{center}$ is the standard center loss. As shown in Figure \ref{fig:taks_2_loss}, we minimize the $L_2$ norm for a same level samples to make intra-class distance smaller and we set a margin $\delta$ between different intensity level samples in feature space. Note that, the margin is depended on the intensity level difference. For example, the distance between \textit{barely engaged} and \textit{engaged} should be smaller than the distance between completely disengaged and engaged. The final RL constitutes $L_{center}$, $L_{rank1}$ and $L_{rank2}$.

\subsection{Bootstrap and Model Ensemble}
The bootstrap method \cite{Efron1979Bootstrap} is a statistical technique to estimate the distribution about dataset by averaging estimates from multiple small data samples.

This approach is called sampling with replacement. The key idea of the method is estimating the true distribution with the empirical distribution. 

To fully make use of the training and validation data, we not only train our model in the official split, but also make three new splits on the dataset, namely $split_1$, $split_2$ and $split_3$. Specifically, we generate three new data splits by utilizing all validation data for training. The new training split consists of all official validation data and some training data. We make the training class balanced as much as possible while we preserve the ratio of training and validation as same as the official one (147:48). We also preserve the subject independence in the new split. Our different splits mainly address some intractable problems if compared with only using official splits. Meanwhile, we can obtain three empirical distributions which can help us estimate the true distribution of the whole dataset. After we get results from various modalities and diverse splits, we ensemble these models by average weights.

	


\section{Experiments}
In this section, we first introduce our implementation details on the \textit{Engagement in the wild} dataset \cite{mustafa2018prediction, dhall2017individual}. Then we show the evaluations of our models trained on different modality features, different splits, and multi-modal ensemble results.


\subsection{Implementation Details}
Based on our winner solution in EmotiW 2018, we design a LSTM layer with 64 hidden states for each modality feature. Then the last state of LSTM is fed into three fully connected layers with size of  {$1024 : 512$, $512 : 128$ and $128 : 1$ respectively}. 

The learning rate is initialized by $0.01$, and multiplied by $0.1$ every 20 epochs. The number of training epoch is set to 60. The weight decay is $5e^{-4}$ and the moment is $0.9$. 

All the implementations are based on \textit{Pytorch}. We preprocess frames by the OpenPose\cite{simon2017hand} and OpenFace toolbox\cite{amos2016openface}. 

We use open source Caffe code to extract C3D features, and we evaluate our approach on both \textit{official split}, \textit{$split_1$}, \textit{$split_2$} and \textit{$split_3$}.

\begin{table}[htp]
	\tiny
	\centering
	\caption{MSE Results on Validation set of \textbf{official split}}
	\label{tab:os}
	\resizebox{\linewidth}{!}{%
		\begin{tabular}{lccc}
			\toprule
			Method & MSE& Normalized MSE \\ \midrule
			1 LSTM + OpenFace  & 0.0853 & 0.0830                      \\
			1 LSTM + OpenPose & \textbf{0.0717} & 0.0739\\
			1 LSTM + C3D & 0.0865 & 0.0897                  \\ 
			 \bottomrule
	\end{tabular}}
\end{table}

\begin{table}[htp]
	\tiny
	\centering
	\caption{MSE Results on Validation set of \textbf{$split_1$}$/$\textbf{2}$/$\textbf{3}}
	\label{tab:ns_1}
	\resizebox{\linewidth}{!}{%
		\begin{tabular}{lccc}
			\toprule
			Method & MSE& Normalized MSE \\ \midrule
			2 LSTM + OpenFace & \textbf{0.0398}/0.0574/0.0721   &   0.0410/0.0583/0.0754\\
			1 LSTM + OpenFace & 0.0428/0.0604/0.0745   &   0.0437/0.0617/0.0767\\
			2 LSTM + OpenPose  & 0.0458/0.0625/0.0725 &  0.0476/0.0636/0.0734       \\
			1 LSTM + OpenPose & 0.0432/0.0621/0.0740 &  0.0454/0.0629/0.0729   \\
			1 LSTM + C3D & 0.0468/0.0653/0.0761 &  0.0459/0.0639/0.0788   \\
			\bottomrule
	\end{tabular}}
\end{table}

\subsection{Experiment Results}
\textbf{Evaluation on official split}
We firstly conduct experiment on official split, and the results are shown in Table \ref{tab:os}. The model using facial, posture and spatio-temporal features get MSE of 0.085, 0.0717 and 0.0865, respectively. The MSE results show that the posture features can contribute more to the engagement intensity regression. Our normalization method is able to improve some model slightly because all the models often generate regression result from 0.2 to 1.0. This is probably because the training samples belonging to 0.0 and 1.0 are too puny.

\textbf{Evaluation on multiple splits}
To fully use the data and the statistical conclusions, we conduct the experiments on our new splits. Shown in Table \ref{tab:ns_1}, it is amazing that our new split effectively reduces the MSE. Our face-based approach on $split_1$ achieves 0.0398 MSE and pose-based one attains best MSE of 0.0428. $Split_2$ and $split_3$ achieve MSE around 0.06 and 0.072, which prove that the MSE performance is sensitive to the data splits.

\textbf{Our submissions}
Eventually, we summarize our 5 submissions as follows. Table \ref{tab:com} shows the corresponding result in the validation and test set. MSE across each intensity level is also provided including \textit{not engaged (NE), barely engaged (BE), engaged (E)} and \textit{strongly engaged (SE)}. All the 5 runs have get competitive results. The first run got the best result in the test dataset. All of the last three runs had reasonable results in the test set.
\begin{enumerate}
	\item OpenFace and OpenPose features in $split_1$.
	\item OpenFace, OpenPose and C3D features in $split_1$.
	\item OpenFace and OpenPose features in $split_1/2/3$.
	\item OpenFace and OpenPose(with 70 face landmarks) features in $split_1$.
	\item (1) + (4)
\end{enumerate}

\begin{table}[htp]
	\centering
	\caption{MSE Results of all models of submissions.}
	\label{tab:com}
	\begin{tabular}{ccccccc}
		\toprule
		Runs & Validation ($\times10^{-4}$) & \multicolumn{4}{c}{Test ($\times10^{-4}$)}                \\ \midrule
		& Overall    & NE & BE & E &  SE & Overall\\
		1    &  \textbf{364}  &  2505  &  \textbf{472}  &  \textbf{153}  &  1627  & \textbf{626}    \\
		2    &  561  &  2781 & 541  & 197  & \textbf{1442}  & 698  \\
		3    &  -  &  2497 & 495  & 177  & 1604  & 643  \\
		4    &  383   & \textbf{2501}  & 484  & 164  & 1615  & 633 \\
		5    &  376  & 2573  & 489  & 158  & 1539 &  638 \\ \bottomrule
	\end{tabular}
\end{table}

\section{Conclusion}
We presented our approach in this paper for the engagement intensity prediction in the Emotion Recognition in the Wild Challenge 2019. We mainly explore the head features and body features with a bootstrap strategy and two novel loss functions. We developed a deep multi-instance learning framework that accepts multiple input features, and evaluated how different modality performs in our framework. OpenFace, OpenPose and C3D features are employed and they compensate for each other. Furthermore, to better estimate the true distribution of the datset, we use the classical bootstrap aggregation method to obtain empirical estimates by resampling the dataset. Experimental results demonstrate the effectiveness of our method, and we eventually win the 3rd place with $0.0626$ MSE.

\begin{acks}
	This work was supported by the National Natural Science Foundation of China (U1613211, 61502152), Shenzhen Basic Research Program (JCYJ20150925163005055, JCYJ20170818164704758), and  International Partnership Program of Chinese Academy of Sciences (172644KYSB20160033,172644KYSB20150019).
\end{acks}

%

\balance
\bibliographystyle{ACM-Reference-Format}
\bibliography{sample-bibliography}

\end{document}